\documentclass[letterpaper,table]{article} %
\usepackage{aaai2027}  %
\usepackage{times}  %
\usepackage{helvet}  %
\usepackage{courier}  %
\usepackage[hyphens]{url}  %
\usepackage{graphicx} %
\usepackage{natbib}  %
\usepackage{caption} %
\usepackage[utf8]{inputenc} %
\usepackage{booktabs}       %
\usepackage{amsfonts}       %
\usepackage{nicefrac}       %
\usepackage{microtype}      %

\usepackage{color}
\usepackage{soul}
\usepackage{multirow}
\usepackage{xcolor}

\newcommand{\thd}[2]{\shortstack{#1\\#2}}

\definecolor{darkred}{rgb}{0.7,0.1,0.1}
\definecolor{darkgreen}{rgb}{0.1,0.7,0.1}
\definecolor{cyan}{rgb}{0.7,0.0,0.7}
\definecolor{dblue}{rgb}{0.2,0.2,0.8}
\definecolor{maroon}{rgb}{0.76,.13,.28}
\definecolor{burntorange}{rgb}{0.81,.33,0}
\definecolor{tealblue}{rgb}{0.212,0.459, 0.533}
\definecolor{myyellow}{rgb}{0.8627451 , 0.75294118, 0.20784314]}

\definecolor{mypink}{rgb}{0.93359375, 0.62109375, 0.83984375}

\definecolor{pp}{rgb}{0.43921569, 0.18823529, 0.62745098}
\definecolor{rr}{rgb}{0.5254902 , 0.00784314, 0.12941176}
\definecolor{bb}{rgb}{0.09019608, 0.23529412, 0.37647059}
\definecolor{yy}{rgb}{0.49803922, 0.3372549 , 0.0}
\definecolor{gg}{rgb}{0.02352941, 0.3372549 , 0.17647059}
\definecolor{mybrown}{rgb}{0.87058824, 0.56078431, 0.01960784}
\definecolor{myblue}{rgb}{0.3372549 , 0.70588235, 0.91372549}
\definecolor{mypurple}{rgb}{0.8, 0.47058824, 0.7372549 }
\definecolor{myorange}{rgb}{0.835, 0.368, 0}
\definecolor{mygreen}{rgb}{0.00784314, 0.61960784, 0.45098039}
\definecolor{mygt}{rgb}{0.0078125 , 0.57421875, 0.40625}
\definecolor{mysp}{rgb}{0.84765625, 0.515625  , 0.0234375}
\definecolor{mycitecolor}{rgb}{0,0.08,0.45}
\definecolor{mygr}{rgb}{0.9607,0.9607,0.9607}
\definecolor{myoo}{rgb}{0.992,0.9176,0.9019}

\definecolor{myrr}{HTML}{AE031A}
\definecolor{mybb}{HTML}{0155B3}

\newcommand{\myparagraph}[1]{\vspace*{2pt}{\bf #1 }}

\usepackage{amsmath,amsfonts,bm}

\def\1{\bm{1}}

\def\sp{space}

\def\vz{{\bm{z}}}

\def\mG{{\bm{G}}}

\def\mS{{\bm{S}}}

\def\mV{{\bm{V}}}

\DeclareMathAlphabet{\mathsfit}{\encodingdefault}{\sfdefault}{m}{sl}
\SetMathAlphabet{\mathsfit}{bold}{\encodingdefault}{\sfdefault}{bx}{n}

\def\gE{{\mathcal{E}}}
\def\gF{{\mathcal{F}}}

\def\gL{{\mathcal{L}}}

\def\gV{{\mathcal{V}}}
\def\gW{{\mathcal{W}}}

\def\sR{{\mathbb{R}}}

\usepackage{symbols}

\usepackage{booktabs}
\usepackage{tabularx}
\usepackage{multirow}       %
\usepackage{algorithm}
\usepackage{algorithmic}
\usepackage{subcaption}
\usepackage{enumitem} %

\title{SpaTime: Streaming Vision-Language Models for Spatio-temporal Reasoning}

\author{
    Hairong Yin\textsuperscript{\rm 1},
    Huangying Zhan\textsuperscript{\rm 2},
    Shin-Fang Chng\textsuperscript{\rm 2},
    Yi Xu\textsuperscript{\rm 2},
    Raymond A. Yeh\textsuperscript{\rm 1}
}
\author{
    Hairong Yin\textsuperscript{\rm 1},
    Huangying Zhan\textsuperscript{\rm 2},
    Shin-Fang Chng\textsuperscript{\rm 2},
    Yi Xu\textsuperscript{\rm 2},
    Raymond A. Yeh\textsuperscript{\rm 1}
}
\affiliations{
    \textsuperscript{\rm 1}Department of Computer Science, Purdue University, USA\\
    \textsuperscript{\rm 2}Goertek Alpha Labs, USA
}

\nocopyright

\begin{document}

\maketitle

\begin{abstract}
  Embodied agents must reason about 3D space while the video is still arriving, answering questions as soon as they have observed enough of the scene. VLMs that incorporate 3D geometric priors achieve strong spatial reasoning, but they operate offline, \ie, the full video must be available before they produce an answer. Streaming VLMs process frames causally and decide for themselves when to respond, yet they lack explicit 3D representations. We present \textbf{SpaTime}, a streaming VLM that fuses causal geometry tokens into the language model at every frame, using only the frames observed so far. To supervise \emph{when} the model answers, we propose a \emph{response-time loss} that maps per-frame response probabilities to a differentiable expected response time and penalizes the distance from the ground-truth frame.
    For evaluation, we construct StreamVSTI-Bench and StreamVSI-Bench, streaming adaptations of VSTI-Bench and VSI-Bench. On StreamVSTI-Bench, SpaTime reaches 49.2\% overall accuracy and reduces the mean response-time error by 66\% relative to the strongest streaming baseline. 
\end{abstract}

\section{Introduction}
\begin{figure}[t]
    \vspace{-0.2cm}
    \centering
    \includegraphics[width=0.98\linewidth]{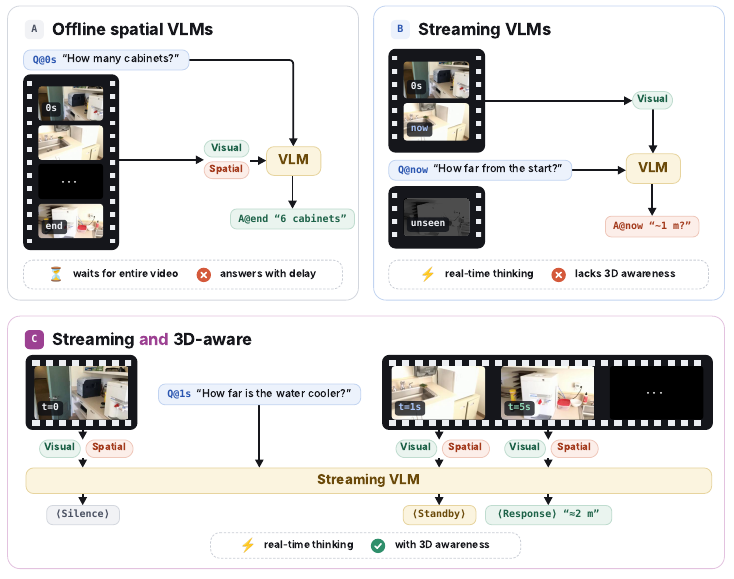}
    \vspace{-0.2cm}
    \caption{Offline VLMs lack real-time answering capability, and general streaming VLMs lack 3D awareness. SpaTime fuses visual and causal 3D geometry tokens to achieve a Streaming VLM for spatial reasoning.}
    \label{fig:teaser}
    \vspace{-0.5cm}
\end{figure}

Spatial reasoning over streaming video is an essential capability for real-world embodied intelligence. Consider a robot navigating a kitchen. When asked ``is the mug to your left or your right?'', the robot must judge the mug's position relative to its own viewpoint, which changes as it moves. When asked ``how far is the mug from where you started?'', it must maintain a geometric history across frames. In both cases, the question may arrive midstream, while the scene is only partially observed. The robot must decide not only \emph{what} to answer, but also \emph{when}. Answering too early means guessing about parts of the scene it has not yet seen, while answering too late stalls the task at hand. Ideally, the robot should answer as soon as it has observed enough information. Deployable embodied and AR systems therefore need a \emph{streaming spatial VLM}, \ie, a VLM that reasons about 3D spatial relationships on a live video stream and decides on its own when to answer (\figref{fig:teaser}).

Recent VLM works mainly study one of these two requirements but not both. Offline VLMs specialized for spatial reasoning, \eg, VG-LLM~\cite{vgllm} and VLM-3R~\cite{vlm3r}, inject 3D geometric priors into the language model and achieve strong performance on spatial benchmarks such as VSI-Bench~\cite{vsibench}. However, every stage of their pipeline, \ie, geometry encoding, feature fusion, and generation, operates over the full frame sequence. They therefore cannot answer a question that arrives mid-stream. Streaming VLMs, \eg, VideoLLM-online~\cite{videollm}, Dispider~\cite{dispider}, and Streamo~\cite{streamo}, take the opposite approach. They process frames causally and emit a state token at each frame, which lets the model decide when to answer. However, they lack any explicit 3D representation and struggle on spatial questions.

The difficulty is that these two requirements pull in opposite directions. Building a coherent 3D representation has meant attending over all input frames at once, as in geometry encoders such as VGGT~\cite{vggt}, which is exactly what causal, frame-by-frame inference forbids. StreamVGGT~\cite{streamVGGT} shows this conflict is resolvable at the representation level, distilling VGGT's global representations into a causal transformer whose key-value cache encodes geometry at constant cost per frame. The encoder is therefore no longer the obstacle, but two pieces above it are still missing. First, the geometry tokens have to reach the language model without reintroducing a dependence on future frames, whereas existing fusion strategies are designed for the offline setting where the full sequence is available. Second, the model has to be supervised on \emph{when} to answer, yet the losses used by streaming VLMs penalize all temporal errors alike, regardless of how early or late the response is.

\begin{figure*}[t]
\centering
\includegraphics[width=0.87\linewidth]{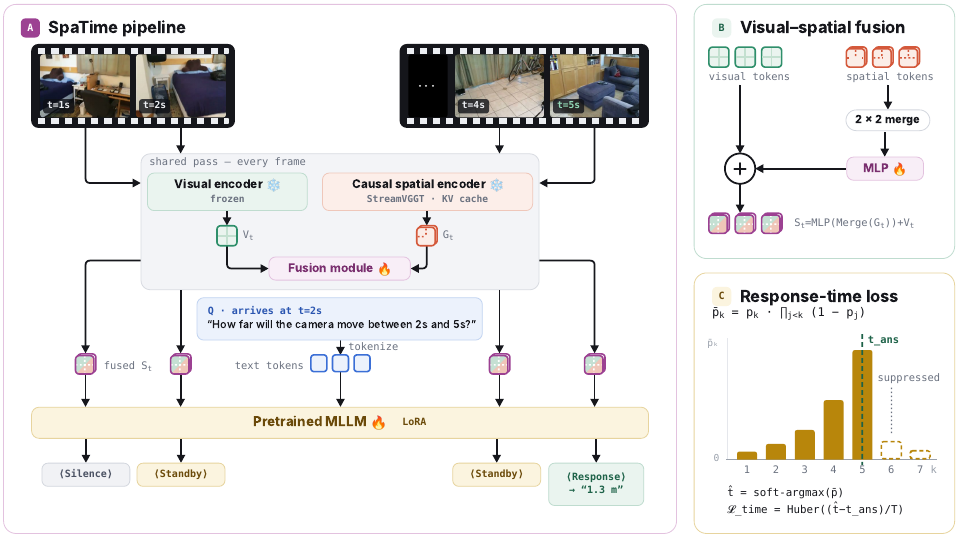}
\caption{SpaTime pipeline. Figure (a) illustrates the overall pipeline. Each frame is encoded by a frozen visual encoder and a frozen causal geometry encoder, then (b) a lightweight projector aligns the geometry tokens with the visual tokens, and fused via element-wise addition. In (c), a response time loss $\gL_\text{time}$ aligns the predicted response time with the ground truth.}
\vspace{-0.2cm}
\label{fig:pipeline}
\end{figure*}

We present \textbf{SpaTime}, a streaming VLM (\figref{fig:pipeline}), to address these challenges. For the first, SpaTime runs a frozen StreamVGGT geometry encoder alongside the VLM's frozen visual encoder and fuses the two at every frame. Following VG-LLM's~\cite{vgllm} fusion strategy, we spatially merge the per-frame geometry tokens, project them through a lightweight MLP, and add them to the visual tokens of the same frame. A LoRA-tuned LLM then reads the fused tokens and predicts a state token at each frame. Because the geometry encoder is causal and the fusion is per-frame, no stage of the pipeline consults a future frame, \ie, it processes only frames $\leq t$ at time $t$, and its inference cost per frame stays constant as the stream grows.

For the second, we propose a \emph{response-time loss} that makes the response time itself trainable. We read per-frame response probabilities off the vocabulary distribution, aggregate them into a differentiable expected response time via soft argmax, and penalize its deviation from the ground-truth frame. The penalty therefore scales with how far the prediction lands from the correct moment, so responding one frame early costs less than responding ten frames early, a distinction that per-frame cross-entropy and focal losses cannot express.

As no spatial reasoning benchmark supports streaming evaluation, we construct StreamVSTI-Bench and StreamVSI-Bench. These benchmarks are streaming adaptations of the popular spatial reasoning benchmarks of VSTI-Bench~\cite{vlm3r} and VSI-Bench~\cite{vsibench} (\secref{sec:benchmarks}). We keep the QA annotations of the original benchmarks and assign each question an arrival time and a ground-truth response time. For evaluation, we consider the standard accuracy performance as well as a measure of responsiveness by $\Delta t$, the mean absolute error between predicted and ground-truth response times. On StreamVSTI-Bench, SpaTime reaches 49.2\% overall accuracy. Notably, it cuts the mean response-time error by 66\% relative to the strongest streaming baseline while achieving 5.1 points higher spatial accuracy.

{\bf\noindent Our contributions are as follows:}
\begin{itemize}[topsep=1pt]
    \item We introduce SpaTime, a streaming VLM that fuses causal 3D geometry tokens into a language model without consulting future frames, at constant cost per frame.
    \item We propose a response-time loss that supervises \emph{when} the model answers, penalizing a predicted response time in proportion to its distance from the correct frame.
    \item We develop a streaming conversion pipeline for spatial QA and use it to build two benchmarks, StreamVSTI-Bench and StreamVSI-Bench. Extensive experiments on both show that SpaTime outperforms streaming baselines and narrows the gap to offline spatial specialists.
\end{itemize}

\section{Related Work}
\myparagraph{Video-language models.} Building on image-language pretraining, video LLMs~\cite{li2023videochat, maaz2024videochatgpt, lin2023videollava, llava-ov, llava-video} extend instruction-tuned dialogue to video. Recent open-source generalists, \eg, Qwen2.5-VL~\cite{qwen2.5vl} and InternVL-2.5~\cite{internvl}, approach proprietary systems~\cite{gpt4o, gemini, claude}, while MovieChat~\cite{song2024moviechat} and MA-LMM~\cite{he2024malmm} compress past frames into memory banks for long video. Note, all of these models process a video as a completed clip and, as VSI-Bench~\cite{vsibench} reports, remain weak at 3D spatial reasoning. We target this challenge under the stricter constraint that frames arrive as a stream.

\myparagraph{Spatial reasoning in vision-language models.} Existing methods
either ground an LLM in explicit 3D input such as point clouds or reconstructed
scenes~\cite{hong20233dllm, chen2024ll3da, huang2024leo}, or reason from 2D
observations alone without global 3D structure~\cite{chen2024spatialvlm, cheng2024spatialrgpt, zhang2025flatland}. Closest to our setting, VSI-Bench~\cite{vsibench} standardized spatial reasoning in egocentric video, prompting methods that inject geometry into video LLMs. Video-3D LLM~\cite{zheng2025video3dllm} adds position-aware encodings, VG-LLM~\cite{vgllm} fuses VGGT~\cite{vggt} tokens into
Qwen2.5-VL~\cite{qwen2.5vl}, and VLM-3R~\cite{vlm3r} fuses CUT3R~\cite{cut3r} features into LLaVA-NeXT-Video~\cite{zhang2024llavanextvideo}. VSTI-Bench~\cite{vlm3r} further extends evaluation to camera and object motion. However, all these methods operate offline and require the full video before answering. Unlike these methods, SpaTime brings explicit geometry to the streaming setting.

\myparagraph{Streaming VLMs.} VideoLLM-online~\cite{videollm} introduced the streaming dialogue paradigm, processing frames sequentially and deciding at each step whether to respond. Flash-VStream~\cite{zhang2024flashvstream} keeps a compressed memory for long streams, Dispider~\cite{dispider} disentangles perception, decision, and reaction, and Streamo~\cite{streamo} adds explicit state tokens with a state-aware focal loss for the silent/responsive class imbalance. Effective as these models are at deciding when to respond, they \textbf{do not} maintain an explicit 3D representation. SpaTime retains their frame-by-frame decision-making and incorporates geometry into it.

\myparagraph{Online 3D reconstruction and causal geometry encoding.}
Online 3D reconstruction spans SLAM systems~\cite{murartal2017orbslam2, teed2021droidslam, zhu2022niceslam} and, more recently, fast feed-forward pointmap regression. DUSt3R~\cite{wang2024dust3r} and MASt3R~\cite{leroy2024mast3r} recover geometry from image pairs without poses, and Spann3R~\cite{wang2024spann3r}, Fast3R~\cite{yang2025fast3r}, and MonST3R~\cite{zhang2025monst3r} extend this incrementally, at scale, and to dynamic scenes. VGGT~\cite{vggt} and CUT3R~\cite{cut3r} predict dense geometry, but rely on global attention over all frames, which makes them inherently offline. MASt3R-SLAM~\cite{murai2025mast3r} and LeanGate~\cite{xiong2026accelerating} target reconstruction and efficiency rather than language-model integration. Most related to this work is StreamVGGT~\cite{streamVGGT}, which distills VGGT into a causal transformer with a KV cache, yielding geometry tokens at constant per-frame cost with minimal quality loss. SpaTime builds directly on it, feeding causal geometry tokens into a streaming VLM for spatial reasoning.

\myparagraph{Concurrent work.} Stream3D-VLM~\cite{stream3dvlm} concurrently develops an online VLM for 3D spatial understanding, likewise building on a frozen StreamVGGT encoder. It fuses geometry through stacked cross-attention and casts response timing as next-token prediction over two decision tokens under a weighted token-level loss. In contrast, SpaTime fuses by additive projection on an exactly aligned token grid, adopts a three-state protocol, and supervises timing with a differentiable response-time loss on the expected response time. Additionally, our introduced benchmarks are streaming conversions of established ones, which keeps accuracy directly comparable with the offline literature.

\section{Preliminaries}
We review the background and notation needed to understand this paper, including the streaming task formulation, Streamo~\cite{streamo} for streaming video-language modeling (\secref{sec:bg_streamo}), and StreamVGGT~\cite{streamVGGT} for causal geometry encoding (\secref{sec:bg_streamvggt}).

\myparagraph{Streaming task formulation.}
Consider a video stream $\gV = \{f_t\}_{t=0}^{T-1}$, where each frame $f_t \in \sR^{3 \times H \times W}$ arrives sequentially. At an arbitrary time step $t_\text{que} \in \{0, \ldots, T{-}1\}$, a user poses a question $Q_\text{txt}$. The model must select a response time $t_\text{ans} \in \{t_\text{que}, \ldots, T{-}1\}$ and produce an answer $A_\text{txt}$. Note, both decisions are made causally, \ie, from the frames $\{f_0, \ldots, f_{t_\text{ans}}\}$ observed up to that point.

\subsection{Streaming Video-Language Modeling}
\label{sec:bg_streamo}
Streamo~\cite{streamo} formulates streaming video understanding as a multi-turn dialogue between a user and an assistant. The video is sampled at a fixed rate, and each frame $f_t$ corresponds to one conversational turn with a timestamp $\langle t\text{s}{-}(t{+}1)\text{s}\rangle$. A visual encoder $\gE_\text{vis}$ maps each frame to a sequence of visual tokens,
\begin{equation}
\label{eq:vis_enc}
  \mV_t = \gE_\text{vis}(f_t),
  \quad \mV_t \in \sR^{N_v \times d},
\end{equation}
where $N_v$ is the number of visual tokens per frame and $d$ is the hidden dimension of the LLM backbone $\gF$. Text inputs are tokenized and embedded by the LLM's embedding layer. At each frame, the visual tokens and any associated text tokens are concatenated and fed to $\gF$.

\myparagraph{State tokens.}
Streamo embeds frame-level response decisions directly into the language model's vocabulary. Specifically, it introduces three special tokens, and the model starts its assistant turn at each frame $t$ by autoregressively generating one of them: (a) $\langle\texttt{Silence}\rangle$ keeps the model silent while it continues to process the incoming frame stream; (b) $\langle\texttt{Standby}\rangle$ signals that relevant visual cues have been identified but more context is still needed; and (c) $\langle\texttt{Response}\rangle$ triggers the generation of the answer text $A_\text{txt}$. The output space at each frame is therefore
\begin{equation}
\small
\label{eq:output_space}
  \hat{o}_t \in \gW
  \cup \bigl\{
    \langle\texttt{Silence}\rangle, \;
    \langle\texttt{Standby}\rangle, \;
    \langle\texttt{Response}\rangle
  \bigr\},
\end{equation}
where $\gW$ denotes the LLM's base vocabulary.

\myparagraph{State-aware focal loss.}
Most frames of a stream are silent, so the class distribution over state tokens is heavily imbalanced and standard cross-entropy training collapses to predicting $\langle\texttt{Silence}\rangle$ at every frame. Streamo addresses this with a state-aware focal loss, which applies inverse-frequency weighting and a focal modulation term at the state-token positions,
\begin{equation}
\label{eq:focal}
  \gL_\text{focal}
  = -\sum_{i} \alpha_{c_i} \,
    (1 - p_i)^{\gamma} \,
    \log p_i,
\end{equation}
where $p_i$ is the predicted probability of the ground-truth token at position $i$, $\alpha_{c_i}$ is a class-frequency balancing weight for the ground-truth class $c_i$, and $\gamma$ is the focal exponent.

\subsection{Causal Geometry Encoding}
\label{sec:bg_streamvggt}

StreamVGGT~\cite{streamVGGT} replaces VGGT's~\cite{vggt} global temporal attention with \emph{causal} temporal attention, so that each frame $t$ attends only to frames $\leq t$,
\begin{equation}
\label{eq:streamvggt}
  \mG_t = \gE_\text{geo}(f_t, \, {\tt KV}_{<t}),
  \quad \mG_t \in \sR^{N_g \times d_g},
\end{equation}
where $\gE_\text{geo}$ denotes the full StreamVGGT pipeline and ${\tt KV}_{<t}$ is the cached key-value state from all previous frames. At inference, only the current frame's tokens are computed, while past frames contribute through the cache. The per-frame cost is constant, \ie, independent of the sequence length $T$.

To compensate for the reduced context under causal attention, StreamVGGT is trained by distillation from a pretrained VGGT teacher, matching the teacher's geometry tokens. This preserves most of VGGT's representational quality while enabling streaming inference.

We use StreamVGGT as a frozen geometry encoder, whose per-frame tokens $\mG_t$ serve as the 3D prior that we fuse into the streaming VLM (\secref{sec:arch}).

\section{Approach}
SpaTime incorporates causal 3D geometry encoding, geometry-visual fusion, and frame-level response control via state tokens in a single streaming architecture, illustrated in \figref{fig:pipeline}. We first describe the geometry-aware streaming architecture (\secref{sec:arch}), then the response-time loss that supervises when the model should respond (\secref{sec:rt_loss}).

\subsection{Geometry-Aware Streaming VLM}
\label{sec:arch}
At each frame $f_t$, the visual encoder $\gE_\text{vis}$ generates visual tokens $\mV_t \in \sR^{N_v \times d}$ (\equref{eq:vis_enc}), and the geometry encoder $\gE_\text{geo}$ generates geometry tokens $\mG_t \in \sR^{N_g \times d_g}$ (\equref{eq:streamvggt}). Both encoders remain frozen throughout training. We update only the fusion projector, the LLM's LoRA parameters, the state-token embeddings, and the LM head.

\myparagraph{Geometry-visual fusion.}
Following VG-LLM~\cite{vgllm}, we fuse geometry tokens into the visual token stream via spatial merging and additive fusion. Specifically, we first merge the geometry token grid with a $2 \times 2$ window, concatenating four adjacent patches into a single vector, and then project the merged tokens to the LLM's hidden dimension $d$:
\begin{equation}
  \label{eq:fusion}
  \mS_t
  = \underbrace{
    \mathrm{MLP}\!\bigl(\mathrm{Merge}_{2\times2}(\mG_t)\bigr)
  }_{\in\;\sR^{N_v \times d}}
  + \;\mV_t,
\end{equation}
where $\mathrm{MLP}: \sR^{4 d_g} \to \sR^{d}$ is a two-layer MLP. Note, the merge brings the geometry grid to the same $N_v$ tokens as the visual grid, so the two are added position by position. We feed the fused tokens $\mS_t$ into the LLM backbone $\gF$ as the visual input for frame $t$.

\myparagraph{State-token streaming.}
The LLM backbone follows Streamo's multi-turn protocol (\secref{sec:bg_streamo}). At each frame $t$, we concatenate the fused tokens $\mS_t$, the timestamp marker, and any pending question texts as the user turn. The model then generates an assistant turn, beginning with one of the three state tokens
$\{\langle\texttt{Silence}\rangle,
  \langle\texttt{Standby}\rangle,
  \langle\texttt{Response}\rangle\}$. If $\langle\texttt{Response}\rangle$ is predicted, answer text follows. A causal attention mask ensures that tokens at frame $t$ attend only to tokens from frames $\leq t$, which preserves strict causality across the pipeline.

\begin{figure*}[t]
    \centering
    \includegraphics[width=0.85\linewidth]{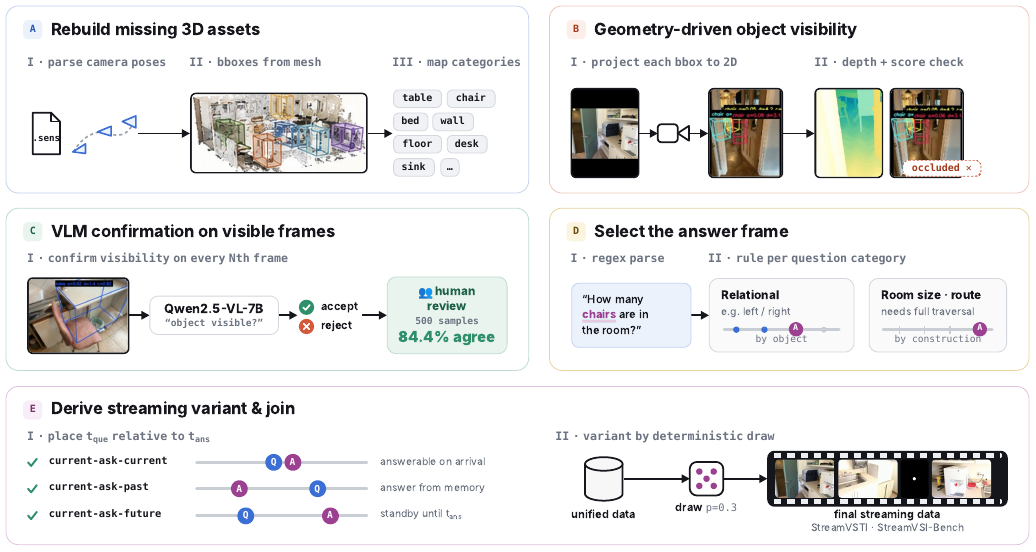}
    \vspace{-0.1cm}
    \caption{Data curation. For each question, a geometry-driven visibility check and a VLM check locate the frames where its objects are visible. Next, a per-type rule selects the answer frame, and a query frame is assigned to create the streaming data.
    }
    \label{fig:benchmark}
    \vspace{-0.25cm}
\end{figure*}

\subsection{Response Time Loss}
\label{sec:rt_loss}
The state-aware focal loss (\equref{eq:focal}) determines \emph{which} state token to predict at each frame, but it treats temporal misalignment uniformly. That is, a model predicting $\langle\texttt{Response}\rangle$ one frame early incurs the same penalty as predicting it ten frames early. We therefore introduce a loss that supplies a differentiable signal for \emph{when} to respond.

\myparagraph{Formulation.}
Let $\{t_1, \ldots, t_T\}$ denote the $T$ state-token positions in a training sample. At each predictor position $t_k - 1$, \ie, the row whose next-token distribution determines the state token at $t_k$, we extract the three-class logit slice corresponding to $\{\langle\texttt{Silence}\rangle, \langle\texttt{Standby}\rangle, \langle\texttt{Response}\rangle\}$:
\begin{equation}
  \label{eq:logit_slice}
  \tilde{\vz}_{t_k}
  = \vz_{t_k - 1}\bigl[\{\langle\texttt{Sil}\rangle, \langle\texttt{Stb}\rangle, \langle\texttt{Resp}\rangle\}\bigr]
  \in \sR^{3},
\end{equation}
and compute a conditional response probability at each state-token frame:
\begin{equation}
  \label{eq:p_resp}
  p_k
  = \mathrm{softmax}\bigl(\tilde{\vz}_{t_k}\bigr)
  [\langle\texttt{Response}\rangle].
\end{equation}

Streaming evaluation protocols only score a model's initial response. To align training with this metric, we model the probability that frame $k$ is the \textit{first} to emit $\langle\texttt{Response}\rangle$:
\begin{equation}
  \label{eq:first_fire}
  \bar{p}_k
  = p_k \prod_{j < k} (1 - p_j),
\end{equation}
which we evaluate in log-space, $\log \bar{p}_k = \log p_k + \sum_{j<k} \log(1 - p_j)$, for numerical stability.
To obtain a differentiable response time prediction, we use the soft-argmax~\cite{luvizon2019human} over $\bar{p}_k$:

\begin{equation}
  \label{eq:soft_argmax}
  \hat{t}_\text{ans}
  = {(\sum_{k=1}^{T} t_k \cdot \bar{p}_k)} \Big/
  {(\sum_{k=1}^{T} \bar{p}_k)},
\end{equation}
with a safe fallback $\hat{t}_\text{ans} = t_T$ when $\sum_k \bar{p}_k$ falls below $10^{-4}$.

The response-time loss then penalizes the length-normalized deviation from the ground-truth response frame $t_\text{ans}$ via a Huber loss:
\begin{equation}
  \label{eq:rt_loss}
  \gL_\text{time}
  = \lambda_\text{time} \cdot
  \mathrm{Huber}_{\delta}\!\left(
  \frac{\hat{t}_\text{ans} - t_\text{ans}}{T}
  \right),
\end{equation}
where $\lambda_\text{time}$ is a scalar weight and $\mathrm{Huber}_{\delta}$ is the Huber loss with $\delta = 0.2$, which limits the effect of outliers.

\myparagraph{Type-weighted response-time loss.}
Not every question requires a timing decision based on visual cues. Among the three temporal variants discussed later in~\secref{sec:benchmarks}, only \emph{current-ask-future} questions require the model to wait. \emph{Current-ask-current} and \emph{current-ask-past} questions are answerable the moment they arrive, so penalizing their response time hurts the state-token posterior for little timing gain.

We therefore introduce a per-type weight to the response-time loss. Let $\tau \in \{\texttt{cur}, \texttt{fut}, \texttt{past}\}$ denote the temporal type of a training sample, read from the per-sample label stored during benchmark construction, and let $w_\tau$ be the corresponding weight. Scaling each sample's loss by $w_\tau$ turns \equref{eq:rt_loss} into
\begin{equation}
  \label{eq:rt_loss_weighted}
  \gL_\text{time}
  = \lambda_\text{time} \cdot  w_\tau \cdot
  \mathrm{Huber}_{\delta}\!\left(
  \frac{\hat{t}_\text{ans} - t_\text{ans}}{T}
  \right).
\end{equation}
The weights emphasize questions that require waiting. We give their specific values in Appendix~\secref{sec:supp_impl}.

\myparagraph{Combined objective.}
The full training loss is: %
\begin{equation}
  \label{eq:total_loss}
  \gL
  = \gL_\text{focal}
  + \gL_\text{time},
\end{equation}
where $\gL_\text{focal}$ is the focal cross-entropy (\equref{eq:focal}) over all supervised assistant-turn tokens and $\gL_\text{time}$ is the type-weighted response-time loss (\equref{eq:rt_loss_weighted}).

\section{Streaming Spatial Reasoning Benchmarks}
\label{sec:benchmarks}

Existing spatial benchmarks, \eg, VSTI-Bench~\cite{vlm3r} and VSI-Bench~\cite{vsibench}, evaluate full videos offline, so they cannot capture \emph{when} an answer becomes viable. We therefore convert them into StreamVSTI-Bench and StreamVSI-Bench with a geometry-grounded pipeline. We preserve the original questions and ground-truth answers built on ScanNet~\cite{dai2017scannet}, ScanNet++~\cite{yeshwanth2023scannet++}, and ARKitScenes~\cite{baruch2021arkitscenes}, which keeps accuracy directly comparable with the offline literature. As shown in~\figref{fig:benchmark}, the pipeline has four stages: {\bf(a)}~rebuilding the 3D assets and a per-frame visibility index (\secref{sec:vis}); {\bf(b)}~confirming recognizability with a VLM (\secref{sec:vlm}); {\bf(c)}~selecting the answer frame $t_\text{ans}$ (\secref{sec:ground}); and {\bf(d)}~deriving the streaming variant (\secref{sec:stream}).

\begin{table*}[t]
\begin{minipage}{0.48\textwidth}
\centering
\renewcommand{\arraystretch}{0.95}
\setlength{\tabcolsep}{3pt}
\resizebox{\linewidth}{!}{
\begin{tabular}{l|c|cc|ccc|c}
\toprule
 & & \multicolumn{2}{c|}{\bf Numerical Answer} & \multicolumn{3}{c|}{\bf Multiple-Choice Answer} & \\
Methods & Avg & \thd{Cam-Obj}{Abs. Dist.} & \thd{Cam.}{Disp.} & \thd{Cam. Move.}{Dir.} & \thd{Obj-Obj}{Rel. Pos.} & \thd{Cam-Obj}{Rel. Dist.} & \thd{$\Delta t$ (s)}{$\downarrow$} \\
\midrule
\rowcolor{blue!5} \multicolumn{8}{l}{\textit{Baseline}} \\
Human Level$^\dagger$ & 77.0 & 51.4 & 46.8 & 95.1 & 97.5 & 94.3 & N/A \\
Chance Level$^\dagger$ & 27.4 & 5.4 & 6.2 & 40.7 & 52.2 & 32.4 & N/A \\
\midrule
\rowcolor{blue!5} \multicolumn{8}{l}{\textit{Proprietary (Offline)}} \\
GPT-4o$^\dagger$ & 38.2 & 29.5 & 23.4 & 37.3 & 58.1 & 42.5 & N/A \\
Gemini-2.5-Pro$^\dagger$ & 42.4 & 29.2 & 8.0 & 30.0 & 84.0 & 60.7 & N/A \\
\midrule
\rowcolor{blue!5} \multicolumn{8}{l}{\textit{Open-source (Offline)}} \\
InternVL2-8B$^\dagger$ & 43.5 & 32.9 & 13.5 & 48.0 & 68.0 & 55.0 & N/A \\
LLaVA-Video-7B$^\dagger$ & 40.0 & 28.2 & 1.8 & 49.8 & 64.7 & 55.6 & N/A \\
LLaVA-OneVision-7B$^\dagger$ & 41.7 & 29.9 & 19.3 & 47.5 & 62.1 & 49.8 & N/A \\
\midrule
\rowcolor{blue!5} \multicolumn{8}{l}{\textit{Offline Spatial VLM}} \\
VG-LLM-8B & 47.1 & 31.3 & 32.5 & 48.2 & 73.4 & 50.0 & N/A \\
VLM-3R-7B$^\dagger$ & 58.8 & 39.4 & 39.6 & 60.6 & 86.5 & 68.6 & N/A \\
\midrule
\rowcolor{blue!5} \multicolumn{8}{l}{\textit{Online Video-LLM}} \\
VideoLLM-online & 48.40 & 33.43 & 32.50 & 45.89 & 68.02 & 69.25 & 29.24 \\
Dispider & 47.25 & 33.40 & 33.43 & 49.68 & 58.78 & 66.91 & 42.59 \\
Streamo & 44.05 & 29.16 & 32.61 & 49.51 & 53.37 & 65.06 & 0.35 \\
\midrule
\rowcolor{blue!5} \multicolumn{8}{l}{\textit{Ours}} \\
Ours & \textbf{49.20} & \textbf{35.33} & \textbf{36.66} & \textbf{46.33} & \textbf{65.41} & \textbf{68.05} & \textbf{0.12} \\
\bottomrule
\end{tabular}
}
\caption{Results on StreamVSTI-Bench. Accuracy is charitable-mode (strict-mode averages are reported separately in Table~\ref{tab:strict}); $\Delta t$ is the mean response-time error in seconds (lower is better), charging a miss the time to video end. Offline models see the full video and are N/A on timing. $^\dagger$: reference numbers from the VLM-3R~\cite{vlm3r} paper.}
\label{tab:vsti}

\end{minipage}
\hfill
\begin{minipage}{0.48\textwidth}
\centering
\renewcommand{\arraystretch}{0.82}
\setlength{\tabcolsep}{2pt}
\resizebox{\linewidth}{!}{
\begin{tabular}{l|c|cccc|cccc|c}
\toprule
 & & \multicolumn{4}{c|}{Numerical Answer} & \multicolumn{4}{c|}{Multiple-Choice Answer} & \\
Methods & Avg & \thd{Obj.}{Count} & \thd{Abs.}{Dist.} & \thd{Obj.}{Size} & \thd{Room}{Size} & \thd{Rel.}{Dist.} & \thd{Rel.}{Dir.} & \thd{Route}{Plan} & \thd{Appr.}{Order} & \thd{$\Delta t$ (s)}{$\downarrow$} \\
\midrule
\rowcolor{blue!5} \multicolumn{11}{l}{\textit{Baseline}} \\
Human Level$^\dagger$ & 79.2 & 94.3 & 47.0 & 60.4 & 45.9 & 94.7 & 95.8 & 95.8 & 100.0 & N/A \\
Chance Level$^\dagger$ & 34.0 & 62.1 & 32.0 & 29.9 & 33.1 & 25.1 & 47.9 & 28.4 & 25.2 & N/A \\
\midrule
\rowcolor{blue!5} \multicolumn{11}{l}{\textit{Proprietary (Offline)}} \\
GPT-4o$^\dagger$ & 34.0 & 46.2 & 5.3 & 43.8 & 38.2 & 37.0 & 41.3 & 31.5 & 28.5 & N/A \\
Gemini-2.5-Pro$^\dagger$ & 51.5 & 43.8 & 34.9 & 64.3 & 42.8 & 61.1 & 47.8 & 45.9 & 71.3 & N/A \\
\midrule
\rowcolor{blue!5} \multicolumn{11}{l}{\textit{Open-source (Offline)}} \\
InternVL2-8B$^\dagger$ & 34.6 & 23.1 & 28.7 & 48.2 & 39.8 & 36.7 & 30.7 & 29.9 & 39.6 & N/A \\
LLaVA-Video-7B$^\dagger$ & 35.6 & 48.5 & 14.0 & 47.8 & 24.2 & 43.5 & 42.4 & 34.0 & 30.6 & N/A \\
LLaVA-OneVision-7B$^\dagger$ & 32.4 & 47.7 & 20.2 & 47.4 & 12.3 & 42.5 & 35.2 & 29.4 & 24.4 & N/A \\
\midrule
\rowcolor{blue!5} \multicolumn{11}{l}{\textit{Offline Spatial VLM}} \\
VG-LLM-8B$^\dagger$ & 50.7 & 67.9 & 37.7 & 58.6 & 62.0 & 46.6 & 40.7 & 32.4 & 59.2 & N/A \\
VLM-3R-7B$^\dagger$ & 60.9 & 70.2 & 49.4 & 69.2 & 67.1 & 65.4 & 80.5 & 45.4 & 40.1 & N/A \\
\midrule
\rowcolor{blue!5} \multicolumn{11}{l}{\textit{Online Video-LLM}} \\
VideoLLM-online & 25.92 & 31.03 & 30.89 & 0.00 & 0.07 & 36.84 & 40.68 & 35.05 & 26.58 & 58.08 \\
Dispider & 24.14 & 27.34 & 32.99 & 0.00 & 0.00 & 34.32 & 40.15 & 28.87 & 34.38 & 58.26 \\
Streamo & 25.23 & 6.7 & 20.6 & 29.3 & 1.6 & 30.3 & 25.54 & 5.8 & 8.3 & 7.73 \\
\midrule
\rowcolor{blue!5} \multicolumn{11}{l}{\textit{Ours}} \\
Ours & \textbf{29.83} & \textbf{38.17} & \textbf{26.58} & \textbf{3.49} & \textbf{0.00} & \textbf{37.70} & \textbf{43.92} & \textbf{38.46} & \textbf{37.71} & \textbf{7.58} \\
\bottomrule

\end{tabular}
}
\caption{Results on StreamVSI-Bench (task types unseen in training). Accuracy is charitable-mode (strict-mode averages in Table~\ref{tab:strict}) and $\Delta t$ is the mean response-time error (s, lower better), as in Table~\ref{tab:vsti}; offline models are N/A on timing. $^\dagger$: reference numbers from VSI-Bench~\cite{vsibench}, VG-LLM~\cite{vgllm}, or VLM-3R~\cite{vlm3r}.}
\label{tab:vsi}
\vspace{-0.1cm}

\end{minipage}
\vspace{-0.25cm}
\end{table*}

\subsection{3D Assets and Geometry-Driven Visibility}
\label{sec:vis}
We first collect 3D asset annotations for each scene. For example, we read per-frame camera poses and intrinsics from the \texttt{.sens} archives in ScanNet. For all datasets, we take object-level oriented bounding boxes (OBBs) from the dataset's instance annotations. We then map each dataset's category labels to a unified question vocabulary, so that every referred object can be grounded to a 3D box.

With these assets in place, we build a per-frame \emph{visibility index} for every object-frame pair. Specifically, we project the eight corners of each object's OBB into the image plane via the camera pose and intrinsics and take the 2D bounding rectangle of the projection. From it we compute an \emph{area ratio}, the fraction of the frame covered by the rectangle clipped to the image, and a \emph{containment ratio}, the fraction of the unclipped rectangle that lies inside the image boundary. Containment rejects a case that a na\"ive area threshold would accept, \ie, a camera close to a large object fills the frame yet observes almost none of it. We consider an object visible when (a)~its containment ratio exceeds a minimum threshold, or its area ratio is large enough that the object dominates the frame even when partially cut off; (b)~its area ratio exceeds a minimum size, rejecting distant specks; (c)~its center lies within a maximum distance from the camera; and (d)~for ARKitScenes and ScanNet++, it is not occluded by intervening geometry in the sensor depth map. The product of the two ratios serves as a per-frame visibility score for ranking views. We report the thresholds in Appendix~\secref{sec:supp_benchmark}.

\begin{table}[t]
    \small
    \setlength{\tabcolsep}{3pt}
    \centering
    \begin{tabular}{c c | c c c}
     \specialrule{.10em}{.01em}{.01em}
     \multicolumn{2}{c|}{Ablations} &
     \multicolumn{3}{c}{Metrics} \\
     Geo. Tokens & RT Loss
     & Strict & Charitable & $\Delta t$\,(s)$\downarrow$ \\
    \hline
      &  & 43.99 & 44.05 & 0.35 \\
     \checkmark &  & 47.93 & 48.37 & 0.18 \\
    \rowcolor{gray!10}
     \checkmark & \checkmark & \textbf{48.80} & \textbf{49.20} & \textbf{0.12} \\
     \specialrule{.10em}{.01em}{.01em}
    \end{tabular}
    \caption{Ablation on StreamVSTI-Bench. Components are added
    cumulatively to a fine-tuned Streamo baseline; the shaded row is our
    final model.}
    \label{tab:ablation}
    \vspace{-.3cm}
\end{table}

\subsection{VLM Confirmation on Visible Frames}
\label{sec:vlm}
Geometric visibility guarantees that a box projects into the frame, not that the object is recognizable there, \eg, motion blur, a doorway edge, or a box that actually sits behind a counter all pass the frustum test. We therefore use a vision-language model to double-check. For each object referenced by a question, we overlay its 3D box as a wireframe on the frames where it is geometrically visible and use Qwen2.5-VL-7B~\cite{qwen2.5vl} to confirm that the object is recognizable. The check is deliberately conservative, \ie, parsing is \emph{fail-open}; only a reply that is an explicit ``no'' overrides geometry, and an object is never rejected for all its frames. This removes early false-positive views without dropping any question.

To validate the resulting visibility labels, we randomly sample 500
object-frame pairs that the pipeline marks as visible and ask human
annotators to judge whether the object is indeed recognizable in the
corresponding frame. We observe a human-pipeline agreement rate of 84.4\%, suggesting that the combined geometric and VLM filtering produces reliable visibility annotations.

\subsection{Answer-Frame Selection}
\label{sec:ground}
To identify the correct answer frame, we first identify the objects from the question texts and match them against the scene's category list from the data annotations. Given the set of referred objects, we select the answer frame $t_\text{ans}$ based on question type. For instance, an object counting question sets $t_\text{ans}$ to the first frame in which all instances of the queried object become visible, since only then can the model produce the correct count. A relational question sets $t_\text{ans}$ to the earliest frame at which all involved objects have appeared. The remaining per-type rules follow similar logic and are listed in Appendix~\secref{sec:supp_benchmark}.

\subsection{Streaming Variants}
\label{sec:stream}
The question and answer responses are categorized into three temporal variants based on the query frame $t_\text{que}$ and the answer frame $t_\text{ans}$:
\begin{itemize}[topsep=1pt]
  \item \emph{Current-ask-current} ($t_\text{que} = t_\text{ans}$): the evidence
    is enough to answer at the moment the question is posed, so the model can answer immediately upon arrival.
  \item \emph{Current-ask-past} ($t_\text{que} > t_\text{ans}$): the question refers to earlier observations and should be answered immediately from memory.
  \item \emph{Current-ask-future} ($t_\text{que} < t_\text{ans}$): the current evidence is not enough to answer, so the model must predict $\langle\texttt{Standby}\rangle$ until the answer frame arrives.
\end{itemize}

Each spatial QA is then randomly assigned to two of its temporal variants. StreamVSTI-Bench comprises $132{,}568$ training and $6{,}042$ test samples over disjoint ScanNet splits, and StreamVSI-Bench contributes $4{,}726$ test samples over $288$ videos drawn from ScanNet, ScanNet++, and ARKitScenes. Further statistics are given in Appendix~\secref{sec:supp_benchmark}.

\section{Experiments}

\begin{table}[t]
    \small
    \setlength{\tabcolsep}{5pt}
    \renewcommand{\arraystretch}{0.95}
    \centering
    \begin{tabular}{l|cc}
    \toprule
    Method & StreamVSTI & StreamVSI \\
    \midrule
    VideoLLM-online & 6.22 & 4.94 \\
    Dispider & 23.39 & 0.63 \\
    Streamo & 43.99 & 21.09 \\
    \midrule
    \rowcolor{gray!10}
    Ours & \textbf{48.80} & \textbf{29.44} \\
    \bottomrule
    \end{tabular}
    \vspace{-0.1cm}
    \caption{Strict-mode average accuracy (\%) for streaming methods. The strict protocol scores only the earliest predicted $\langle\texttt{Response}\rangle$ and counts a missing response as zero, penalizing timing failures directly. Offline models do not respond and are omitted.}
    \label{tab:strict}
    \vspace{-0.45cm}
\end{table}


\myparagraph{Evaluation benchmarks.}
We evaluate on the two streaming benchmarks of~\secref{sec:benchmarks}: (a)~StreamVSTI-Bench, an in-distribution evaluation, and (b)~StreamVSI-Bench, whose task types are unseen in training. All models are finetuned only on the StreamVSTI-Bench training split.

\myparagraph{Metrics.} Following VSI-Bench~\cite{vsibench}, we report accuracy separately for numerical tasks and multiple-choice tasks under two evaluation modes. The \emph{strict} mode scores only the model's earliest predicted $\langle\texttt{Response}\rangle$ and counts a sample that never emits one as zero. The \emph{charitable} mode instead force-decodes an answer at the final frame when no $\langle\texttt{Response}\rangle$ was emitted, isolating spatial reasoning quality from response-timing failures. For streaming models, we also report $\Delta t$, the mean absolute error in seconds between the predicted and ground-truth response times.

\myparagraph{Baselines.} We compare against three baseline categories. Offline general VLMs: InternVL2-8B~\cite{internvl2}, LLaVA-OneVision-7B~\cite{llava-ov}, LLaVA-Video-7B~\cite{llava-video}, GPT-4o~\cite{gpt4o}, and the Gemini-2.5~\cite{gemini25} series. Offline spatial VLMs: VG-LLM~\cite{vgllm} and VLM-3R~\cite{vlm3r}, which run on the full video, so $\Delta t$ does not apply. Streaming VLMs: Streamo~\cite{streamo}, VideoLLM-online~\cite{videollm}, and Dispider~\cite{dispider}, all fine-tuned on our training data and evaluated in streaming mode. Offline numbers are from prior work where available.

\myparagraph{Implementation details.}
SpaTime is fine-tuned with LoRA on the StreamVSTI-Bench training split. The visual and geometry encoders remain frozen (\secref{sec:arch}). Additional implementation details are documented in Appendix~\secref{sec:supp_impl}.

\subsection{Results on StreamVSTI-Bench}
\label{sec:vsti_results}

\tabref{tab:vsti} presents charitable-mode results on StreamVSTI-Bench. Among streaming models, SpaTime achieves the highest spatial accuracy under both modes, beating the strongest streaming baseline by 4.8 points (strict) and 0.8 (charitable) on average. The gap is starker under the strict protocol (\tabref{tab:strict}), where baselines that answer late or never collapse, whereas SpaTime stays close to its charitable accuracy. In addition, SpaTime's $\Delta t$ of 0.12\,s versus 0.35\,s for the best streaming baseline shows its response times closely track the ground truth, and it comes within 9.6 points of VLM-3R while decoding causally.

\subsection{Generalization to StreamVSI-Bench}
\label{sec:vsi_results}

We further evaluate generalization on StreamVSI-Bench, whose task types are absent from training. \tabref{tab:vsi} reports the full per-category breakdown. Since offline models cannot run under the streaming protocol, we cite their published VSI-Bench~\cite{vsibench} results as an offline reference, marked $^\dagger$. SpaTime outperforms the streaming baselines by 4.7 points on average, showing that the geometry priors generalize beyond the trained task types.

The experiment also reveals a clear shortcoming: SpaTime scores $0.0$ on room-size and $3.5$ on object-size estimation. Both categories require predicting an absolute metric extent, yet their question templates are absent from the camera- and object-relational splits of StreamVSTI-Bench used for training, so the model
never encounters their answer form during training. Other streaming baselines exhibit the same failure on these categories, confirming that the cause is insufficient training coverage rather than a fundamental geometric limitation. Because our pipeline can synthesize size-estimation questions automatically, augmenting the training mixture with such samples is a promising direction for future work.

\subsection{Ablation Study}
\label{sec:ablation}

We ablate SpaTime's components in \tabref{tab:ablation} by progressively adding
each module to a fine-tuned Streamo baseline. Geometry tokens from StreamVGGT
yield the largest gain, improving accuracy by 3.9 points (strict) and 4.3
(charitable) while cutting $\Delta t$ from 0.35\,s to 0.18\,s, confirming that
explicit 3D priors address the core deficit of streaming VLMs on spatial tasks.
Adding the response-time loss further raises charitable accuracy to 49.20 and
reduces $\Delta t$ to 0.12\,s.
By construction, the response-time loss supervises \emph{current-ask-future} questions the most, whose answers depend on frames arriving
after the query. \tabref{tab:rtl} shows that while the overall
$\Delta t$ decreases modestly (0.18\,s to 0.12\,s), the error on future
questions drops from 0.52\,s to 0.35\,s. This shows the benefit in the hard timing cases that the loss is designed to address.

\begin{table}[t]
    \small
    \setlength{\tabcolsep}{5pt}
    \renewcommand{\arraystretch}{0.95}
    \centering
    \begin{tabular}{c c | c c}
     \specialrule{.10em}{.01em}{.01em}
     \multicolumn{2}{c|}{Ablations} & \multicolumn{2}{c}{$\Delta t$\,(s)$\downarrow$} \\
     Geo. Tokens & RT Loss & All & Future-only \\
    \hline
     \checkmark &            & 0.18 & 0.52 \\
     \checkmark & \checkmark & \textbf{0.12} & \textbf{0.35} \\
     \specialrule{.10em}{.01em}{.01em}
    \end{tabular}
    \vspace{-0.1cm}
    \caption{The response time loss targets future questions. Mean
    response-time error $\Delta t$\,(s, lower is better) over all questions
    versus \emph{current-ask-future} questions only, whose answers depend on
    frames that arrive after the query. Adding the response time loss (row~2)
    cuts the error most on future questions, where the model must decide how
    long to wait.}
    \label{tab:rtl}
    \vspace{-0.45cm}
\end{table}

\vspace{0.05cm}
\section{Conclusion}
\vspace{0.05cm}
We propose SpaTime, a streaming VLM that brings explicit 3D geometry into causal video reasoning. SpaTime fuses causal geometry tokens from a frozen StreamVGGT encoder into a streaming LLM at every frame, and a response-time loss supervises \emph{when} the model answers. To train and evaluate in this setting, we construct StreamVSTI-Bench and StreamVSI-Bench, streaming versions of established spatial QA benchmarks. On StreamVSTI-Bench, SpaTime reaches 49.2\% overall accuracy and cuts the mean response-time error from 0.35\,s to 0.12\,s relative to the strongest streaming baseline. We hope SpaTime and these two benchmarks encourage future work on VLMs that reason about 3D space and respond promptly as video frames continue to arrive.

\bibliography{ref}

\clearpage
\appendix

\setcounter{section}{0}
\renewcommand{\thetable}{A\arabic{table}}
\setcounter{table}{0}
\setcounter{figure}{0}
\setcounter{equation}{0}
\renewcommand{\thetable}{A\arabic{table}}
\renewcommand\thefigure{A\arabic{figure}}
\renewcommand{\theHtable}{A.Tab.\arabic{table}}%
\renewcommand{\theHfigure}{A.Abb.\arabic{figure}}%
\renewcommand\theequation{A\arabic{equation}}
\renewcommand{\theHequation}{A.Abb.\arabic{equation}}%

\begin{figure*}[t]
    \centering
    \includegraphics[width=0.9\linewidth]{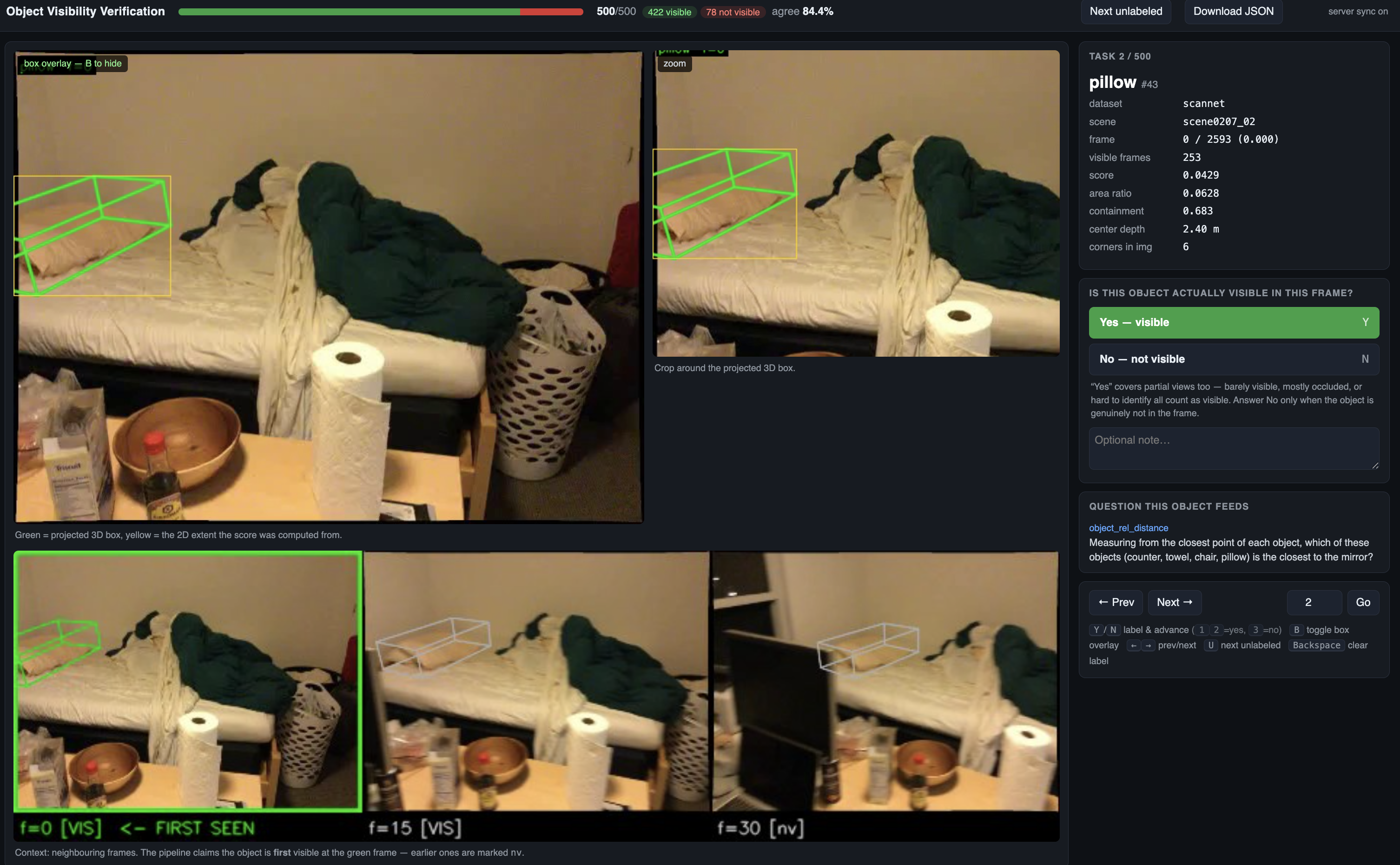}
    \caption{Interface for the human verification study of the visibility annotations. For each of the $500$ sampled object--frame pairs, annotators see the projected 3D box overlay, a zoomed crop, and neighboring frames around the pipeline's first-seen claim, and judge whether the object is genuinely recognizable; the resulting human--pipeline agreement is $84.4\%$.}
    \label{fig:human_review}
\end{figure*}

\section{Appendix}
The appendix is organized as follows:
\begin{itemize}
    \item In Appendix~\secref{sec:supp_impl}, we provide additional implementation details, the full hyperparameter configuration, and the evaluation protocol.
    \item In Appendix~\secref{sec:supp_benchmark}, we provide additional details on the construction of StreamVSTI-Bench and StreamVSI-Bench, including the human verification study, the temporal-channel timing rules, and the full QA template set.
    \item In Appendix~\secref{sec:supp_qual}, we provide qualitative results, including a per-channel timing analysis and round-by-round streaming traces.
\end{itemize}

\section{Implementation Details}
\label{sec:supp_impl}

\myparagraph{Architecture and data.}
SpaTime is built on Qwen2.5-VL-7B-Instruct~\cite{qwen2.5vl} with a frozen
StreamVGGT~\cite{streamVGGT} causal geometry encoder. 
The trainable geometry projector applies RMSNorm, a $2\times2$ spatial merge,
and a two-layer MLP, producing $324$ tokens per frame that are added to the
Qwen2.5-VL visual tokens (\equref{eq:fusion}); the ViT and vision--language
aligner remain frozen. The language model is adapted with LoRA on all linear
projections, while \texttt{embed\_tokens} and \texttt{lm\_head} are trained in
full so that the three state tokens can be learned, yielding $15.0\%$ trainable
parameters. Training uses the $132{,}568$-sample StreamVSTI-Bench training set,
of which $1\%$ is held out for validation.

\myparagraph{Training and optimization.}
The training objective is the combined loss of \equref{eq:total_loss}: a
state-aware focal cross-entropy plus the type-weighted response time loss of
\equref{eq:rt_loss_weighted}, for which we set $w_\texttt{fut} = 1$ and
$w_\texttt{cur} = w_\texttt{past} = 0$ with $\lambda_\text{time}=3.0$, so that
timing supervision is applied only to current-ask-future samples ($41{,}196$ of
$132{,}568$), whose answers must wait for evidence that has not yet appeared.
We optimize with AdamW under a cosine schedule in \texttt{bf16} on $4$ NVIDIA
GH200 GPUs, validating every $50$ steps with early
stopping. The
full hyperparameter configuration is listed in \tabref{tab:hparams}.

\begin{table}[t]
    \small
    \setlength{\tabcolsep}{5pt}
    \renewcommand{\arraystretch}{1.0}
    \centering
    \begin{tabular}{l l}
    \toprule
    Hyperparameter & Value \\
    \midrule
    LoRA rank / $\alpha$ / dropout & $128$ / $256$ / $0.05$ \\
    LoRA targets & \texttt{q,k,v,o,gate,up,down} \\
    Fully trained modules & \texttt{embed\_tokens}, \texttt{lm\_head} \\
    Trainable parameters & $1{,}461.2$M ($15.0\%$) \\
    Base / projector learning rate & $1\times10^{-5}$ / $5\times10^{-4}$ \\
    Optimizer & AdamW, $\beta=(0.9,0.95)$ \\
    Weight decay / grad.\ clip & $0.1$ / $1.0$ \\
    Schedule & cosine, $5\%$ warm-up \\
    Precision & \texttt{bf16} (projector \texttt{fp32}) \\
    Batch size / grad.\ accum. & $1$ per device / $8$ \\
    Max sequence length & $20{,}480$ tokens \\
    Frame rate / geometry input & $1$\,fps / $518\times518$ \\
    Visual pixel budget per frame & $3{,}136$--$100{,}352$ \\
    $\lambda_\text{time}$ / Huber $\delta$ / focal $\gamma$ & $3.0$ / $0.2$ / $2$ \\
    Seed & $42$ \\
    Eval.\ / checkpoint interval & every $50$ steps \\
    Early-stop patience & $3$ evaluations \\
    Hardware (train / eval) & $4\times$ GH200 / RTX A6000 \\
    \bottomrule
    \end{tabular}
    \caption{Training hyperparameters of SpaTime.}
    \label{tab:hparams}
\end{table}

\myparagraph{Evaluation protocol.}
All evaluations run on an NVIDIA RTX A6000 streaming at $1$\,fps. Multiple-choice accuracy is
exact match on the first token of the prediction; numerical questions follow
VSI-Bench~\cite{vsibench} mean relative accuracy (MRA), the fraction of
thresholds $\theta \in \{0.50, 0.55, \dots, 0.95\}$ whose relative error
$|\hat{y} - y| / |y|$ stays within $1 - \theta$. The overall score is the
unweighted mean of the two, under the strict and charitable modes defined in
the main text.

\begin{figure*}[t]
\footnotesize
\begin{verbatim}
Sample: camera_movement_direction, current-ask-future, scene0616_00

Q (asked at round 39): What will be the primary consistent direction of the camera's movement
                       relative to its orientation from 27.68s to 83.04s?
                       Options: A. Left   B. Backward   C. Forward

Rounds  0-38 : </Silence>      (question not yet asked)
Rounds 39-82 : </Standby>      (44 consecutive rounds -- queried interval in progress)
Round     83 : </Response> C   (ground truth: round 83, answer C -- correct)
Rounds 84-86 : </Silence>

Round census: 42 </Silence>, 44 </Standby>, 1 </Response>.
\end{verbatim}
\caption{Round-by-round streaming trace of a current-ask-future sample. The model stays silent before the question, holds $\langle\texttt{Standby}\rangle$ for the $44$ rounds during which the queried interval is still in progress, and fires $\langle\texttt{Response}\rangle$ with the correct answer on exactly the ground-truth round.}
\label{fig:qual_trace}
\end{figure*}

\begin{table}[t]
    \small
    \setlength{\tabcolsep}{6pt}
    \centering
    \begin{tabular}{l c c c}
    \toprule
    Split & Future & Current & Past \\
    \midrule
    Train ($132{,}568$) & $41{,}196$ & $22{,}268$ & $69{,}104$ \\
    Test ($6{,}042$)$^{*}$ & $2{,}076$ & $2{,}075$ & $1{,}891$ \\
    \bottomrule
    \end{tabular}
    \caption{Temporal-channel distribution of StreamVSTI-Bench.}
    \label{tab:channel_dist}
\end{table}

\begin{figure*}[t]
\footnotesize
\begin{verbatim}
[camera_displacement -- numeric]
 future : How far (in meters) will the camera move between {t_start}s and {t_end}s? {postfix}
 current: What is the camera displacement (in meters) from {t_start}s to {t_end}s? {postfix}
 past   : Looking back, how far did the camera move between {t_start}s and {t_end}s in
          meters? {postfix}

[camera_movement_direction -- multiple choice]
 future : What will be the primary consistent direction of the camera's movement relative to its
          orientation from {t_start}s to {t_end}s? {options}
 current: What is the primary consistent direction of the camera's movement relative to its
          orientation from {t_start}s to {t_end}s? {options}
 past   : Looking back, what was the primary consistent direction of the camera's movement relative
          to its orientation from {t_start}s to {t_end}s? {options}

[camera_obj_abs_dist -- numeric]
 future : What is the approximate distance (in meters) between the camera (or the person filming)
          and the nearest point of the {object} at {t_target}s? {postfix}
 current: At current time, what is the approximate distance (in meters) between the camera (or the
          person filming) and the nearest point of the {object}? {postfix}
 past   : Looking back, what is the approximate distance (in meters) between the camera (or the
          person filming) and the nearest point of the {object} at {t_target}s? {postfix}

[camera_obj_rel_dist_v1/v2/v3 -- multiple choice, 2/3/4 candidate objects]
 future : Measuring from the closest point of each object at {t_target}s, which of these objects
          {objects} will be the closest to the camera?\n{options}
 current: Measuring from the closest point of each object at current time, which of these objects
          {objects} will be the closest to the camera?\n{options}
 past   : Looking back, measuring from the closest point of each object at {t_target}s, which of
          these objects {objects} will be the closest to the camera?\n{options}

[obj_obj_relative_pos_lr/ud/nf -- multiple choice]
 future : At {t_target}s, will {obj_A} be to the {axis} relative to {obj_B}?\n{options}
 current: At current time, relative to {obj_B}, is {obj_A} to the {axis}?\n{options}
 past   : Looking back, at {t_target}s, relative to {obj_B}, is {obj_A} to the {axis}?\n{options}

postfix = "\nPlease answer the question using a single word or phrase."
{axis} in [Left/Right], [Up/Down], [Near/Far]
All multiple-choice questions end with: "Answer with the option's letter from the given choices
directly."
\end{verbatim}
\caption{The full QA template set of StreamVSTI-Bench: 5 task families $\times$ 3 temporal channels. The template strings are identical between the training-set and test-set generators.}
\label{fig:qa_templates}
\end{figure*}

\begin{figure*}[t]
\footnotesize
\begin{verbatim}
You are a helpful assistant specializing in streaming video analysis.
You will receive input frame by frame, each labeled with absolute time intervals
in the exact format <Xs-Ys> (e.g., <0s-1s>). Follow these rules precisely:

1. Use </Silence> when:
   - No relevant event has started, OR
   - The current input is irrelevant to the given question.

2. Use </Standby> when:
   - An event is in progress but has not yet completed, OR
   - The current input is relevant but the question cannot yet be answered.

3. Use </Response> only when:
   - An event has fully concluded, OR
   - The available information is sufficient to fully answer the question.
   Provide a complete description at this point.

Do not provide partial answers or speculate beyond the given information.
Whenever you deliver an answer, begin with </Response>.
\end{verbatim}
\caption{The streaming system prompt, reproduced verbatim. The literal token strings \texttt{</Silence>}, \texttt{</Standby>}, and \texttt{</Response>} are rendered as $\langle\texttt{Silence}\rangle$, $\langle\texttt{Standby}\rangle$, and $\langle\texttt{Response}\rangle$ in the main text. Each user turn is \texttt{<Xs-Ys>} followed by the frame; the question text is prepended to the user turn of the asking round, and the assistant emits exactly one state token per round.}
\label{fig:system_prompt}
\end{figure*}

\begin{table}[t]
    \small
    \setlength{\tabcolsep}{4.5pt}
    \renewcommand{\arraystretch}{1.0}
    \centering
    \begin{tabular}{l l c c c c}
    \toprule
    Channel & Bucket & $n$ & Acc.\ (\%) & Exact round & $\Delta$round \\
    \midrule
    Current & MC & $1{,}440$ & $62.50$ & $99.1\%$ & $+0.001$ \\
    Current & Num. & $635$ & $35.04$ & $98.7\%$ & $+0.005$ \\
    Future & MC & $1{,}512$ & $59.92$ & $60.4\%$ & $-0.371$ \\
    Future & Num. & $564$ & $35.32$ & $74.3\%$ & $-0.242$ \\
    Past & MC & $1{,}346$ & $63.45$ & $100.0\%$ & $0$ \\
    Past & Num. & $545$ & $36.90$ & $99.8\%$ & $0$ \\
    \bottomrule
    \end{tabular}
    \caption{Per-temporal-channel breakdown on StreamVSTI-Bench. ``Exact round'' is the fraction of samples whose $\langle\texttt{Response}\rangle$ lands on precisely the ground-truth round; $\Delta$round is the mean of $t_\text{pred} - t_\text{gt}$, so negative values mean the model answers early.}
    \label{tab:qual_channel}
\end{table}

\section{Benchmark Construction Details}
\label{sec:supp_benchmark}

\myparagraph{Dataset statistics.}
StreamVSTI-Bench comprises $132{,}568$ training samples over $1{,}201$ ScanNet scenes and a held-out test set of $6{,}042$ samples over $312$ scenes, with training and test drawn from disjoint source splits so that no test scene is seen during fine-tuning. Each sample averages $71.6$ frames, i.e., roughly $72$ seconds of streaming video at $1$\,fps. StreamVSI-Bench contains $4{,}726$ test samples over $288$ videos ($88$ ScanNet, $150$ ARKitScenes, $50$ ScanNet++), and preserves the full task category of the original VSI-Bench.

The construction pipeline is summarized in \secref{sec:benchmarks}, and we give the thresholds and per-type rules below.

\myparagraph{Visibility index.}
An object-frame pair is accepted when the box's containment is at least $0.2$ or its area ratio at least $0.05$, its area ratio exceeds $0.002$, and its center depth is within $8$\,m. Where a per-frame depth map is available (ARKitScenes, ScanNet++), a box is additionally rejected when a majority of nine probe points (the box center and its eight corners) read more than $0.3$\,m behind the sensor surface; ScanNet has no aligned depth and runs frustum-only. On the retained frames, Qwen2.5-VL-7B confirms recognizability by sampling every $15$th visible frame (capped at $60$ per object) with fail-open parsing.

\myparagraph{Human verification.}
To validate the visibility labels produced by the combined geometric and VLM filtering, we randomly sample $500$ object--frame pairs that the pipeline marks as visible and ask human annotators to judge whether the object is indeed recognizable in the corresponding frame. \figref{fig:human_review} shows the annotation interface where each task presents the full frame with the projected 3D box, a crop around the box, and the neighboring frames surrounding the pipeline's first-seen frame, together with the question that the object feeds. 

\myparagraph{Answer-frame rules.}
After a question's objects are resolved, the answer frame $t_\text{ans}$ follows one of four rules:
\begin{itemize}
    \item \emph{Counting} takes the last of the per-instance first-seen frames.
    \item \emph{Relative/absolute distance}, \emph{relative direction}, and \emph{appearance order} take the last of the per-category first-seen frames.
    \item \emph{Size estimation} takes the single best-framed view, which is scored by the argmax of area ratio $\times$ containment.
    \item \emph{Room size} and \emph{route planning} default to the final frame.
\end{itemize}
If a required object is never visible, the question is left ungrounded.

\myparagraph{Streaming split.}
To form the streaming variant of StreamVSI-Bench, a question becomes \emph{current-ask-future} with probability $0.3$ when its answer frame lies beyond frame $5$; its query frame $t_\text{que}$ is then drawn uniformly from $[0,\,t_\text{ans}-1]$ and its expected response before $t_\text{ans}$ is $\langle\texttt{Standby}\rangle$. Otherwise, $t_\text{que} = t_\text{ans}$ and the question is answerable on arrival.

\myparagraph{Temporal channels on StreamVSTI-Bench.}
For StreamVSTI-Bench, each sample's query frame $t_\text{que}$ and response frame $t_\text{ans}$ are placed relative to the end of the queried interval, denoted $t_\text{target}$, according to its temporal channel:
\begin{itemize}
    \item \emph{current-ask-future}: $t_\text{que}$ is the queried interval's start, jittered by $\pm5$ frames, and $t_\text{ans} = t_\text{target}$, so the model must wait and decide \emph{when} to answer;
    \item \emph{current-ask-current}: $t_\text{que} = t_\text{ans} = t_\text{target}$, so the model answers immediately;
    \item \emph{current-ask-past}: $t_\text{que} = t_\text{target} + \mathcal{U}\{1..5\}$ frames and $t_\text{ans} = t_\text{que}$, so the model answers immediately from its memory of earlier frames.
\end{itemize}
\tabref{tab:channel_dist} gives the resulting channel distribution. \figref{fig:qa_templates} lists the full QA template set, covering the five task families across the three temporal channels, and \figref{fig:system_prompt} shows the streaming system prompt that instructs the model on when to emit each of the three state tokens.

\section{Qualitative Results}
\label{sec:supp_qual}

\myparagraph{Where the timing difficulty lives.}
\tabref{tab:qual_channel} breaks the StreamVSTI-Bench results down by temporal channel. Current and past questions are answerable on the asking round by construction, and the model behaves accordingly where it emits $\langle\texttt{Response}\rangle$ on exactly the ground-truth round in $98.7$--$100\%$ of these samples. Essentially all timing difficulty is therefore concentrated in the future channel, where the model must decide how long to wait; even there it hits the exact ground-truth round $60.4\%$ (multiple-choice) to $74.3\%$ (numerical) of the time, and when it misses the round it gets \emph{early} rather than late (mean signed error $-0.37$ and $-0.24$ rounds).

Accuracy itself is nearly uniform across channels ($59.9$--$63.5\%$ multiple-choice, $35.0$--$36.9\%$ numerical), so waiting does not degrade answer quality. This concentration of timing difficulty in the future channel is the empirical justification for gating the response time loss to current-ask-future samples (\secref{sec:rt_loss}).

\myparagraph{Streaming traces.}
\figref{fig:qual_trace} shows a full round-by-round trace of a current-ask-future sample, and \tabref{tab:qual_examples} collects further examples across all five task families and all three temporal channels, listing the ground-truth and predicted response rounds side by side. In each case, the model emits $\langle\texttt{Response}\rangle$ on exactly the ground-truth round with the correct answer, after holding $\langle\texttt{Standby}\rangle$ for up to $25$ rounds on future questions and recalling up to $\approx97$\,s of history on past questions.

\begin{table*}[t]
    \small
    \setlength{\tabcolsep}{5pt}
    \renewcommand{\arraystretch}{1.1}
    \centering
    \begin{tabular}{l l p{7.6cm} c c}
    \toprule
    Task / channel & Scene & Question (abridged) & GT (round, ans.) & Pred.\ (round, ans.) \\
    \midrule
    Rel.\ pos.\ (lr) / future & \texttt{scene0700\_02} & At 33.40s, will telephone be to the [Left/Right] relative to keyboard? (asked at round 14) & (34, A) & (34, A) \\
    Rel.\ dist.\ (v3) / future & \texttt{scene0580\_00} & Measuring from the closest point of each object at 43.42s, which of (nightstand, bed) will be closest to the camera? (asked at round 24) & (44, B) & (44, B) \\
    Displacement / future & \texttt{scene0664\_00} & How far (in meters) will the camera move between 13.12s and 39.35s? (asked at round 14) & (39, 0.5) & (39, 0.5) \\
    Rel.\ pos.\ (ud) / current & \texttt{scene0231\_02} & At current time, relative to backpack, is window to the [Up/Down]? & (4, A) & (4, A) \\
    Rel.\ dist.\ (v2) / current & \texttt{scene0580\_01} & Measuring from the closest point of each object at current time, which of (backpack, table, bed) is closest to the camera? & (39, C) & (39, C) \\
    Movement dir.\ / past & \texttt{scene0050\_00} & Looking back, what was the primary consistent direction of the camera's movement from 12.11s to 84.80s? & (109, C) & (109, C) \\
    Rel.\ pos.\ (nf) / past & \texttt{scene0050\_02} & Looking back, at 153.03s, relative to camera, is backpack to the [Near/Far]? & (170, B) & (170, B) \\
    \bottomrule
    \end{tabular}
    \caption{Qualitative examples across task families and temporal channels on StreamVSTI-Bench. For every example, the predicted $\langle\texttt{Response}\rangle$ round matches the ground-truth round exactly and the answer is correct; future-channel examples require holding $\langle\texttt{Standby}\rangle$ for $20$--$25$ rounds before responding.}
    \label{tab:qual_examples}
\end{table*}

\end{document}